\documentclass{article}

\usepackage{color, colortbl}

\usepackage[preprint]{neurips_2024}

\usepackage[utf8]{inputenc} 
\usepackage[T1]{fontenc}    
\usepackage{hyperref}       
\usepackage{url}            
\usepackage{booktabs}       
\usepackage{amsfonts}       
\usepackage{nicefrac}       
\usepackage{microtype}      
\usepackage{xcolor}         
\usepackage{graphicx}       
\usepackage{amsmath}       
\usepackage{geometry}
\usepackage{array}
\usepackage{longtable}
\usepackage{footnote}
\makesavenoteenv{longtable}
\usepackage{comment}
\usepackage{pdfpages}
\usepackage{svg}
\usepackage{pifont}

\definecolor{darkgreen}{rgb}{0.05, 0.5, 0.06}
\definecolor{alizarin}{rgb}{0.82, 0.1, 0.26}

\title{Linguini\protect\includegraphics[height=1em]{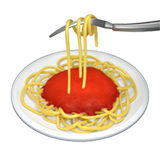}: A benchmark for language-agnostic linguistic reasoning}

\author{Eduardo Sánchez$^{\text{$\star$ $\dag$}}$ \quad Belen Alastruey$^{\text{$\star$}}$ \quad Christophe Ropers$^{\text{$\star$}}$ \\ 
\bf Pontus Stenetorp$^{\text{$\dag$}}$ \quad Mikel Artetxe$^{\text{$\ddag$}}$ \quad Marta R. Costa-jussà$^{\text{$\star$}}$ \\
$^{\text{$\star$}}$Meta \quad
$^{\text{$\dag$}}$University College London \\
$^{\text{$\ddag$}}$University of the Basque Country (UPV/EHU) \\
\texttt{\{eduardosanchez, alastruey, chrisropers, costajussa\}@meta.com} \\
\texttt{p.stenetorp@cs.ucl.ac.uk} \quad \texttt{mikel.artetxe@ehu.eus} \\
}

\begin{document}

\maketitle

\begin{abstract}
We propose a new benchmark to measure a language model's linguistic reasoning skills without relying on pre-existing language-specific knowledge. The test covers 894 questions grouped in 160 problems across 75 (mostly) extremely low-resource languages, extracted from the International Linguistic Olympiad corpus. To attain high accuracy on this benchmark, models don't need previous knowledge of the tested language, as all the information needed to solve the linguistic puzzle is presented in the context. We find that, while all analyzed models rank below 25\% accuracy, there is a significant gap between open and closed models,  with the best-performing proprietary model at 24.05\% and the best-performing open model at 8.84\%. 
\end{abstract}

\section{Introduction}
\label{sec:intro}

Recently, language models have shown impressive multilingual skills \citep{xu2024survey}, achieving state of the art results in several tasks, such as machine translation \citep{openai2024gpt4}, bilingual lexicon induction \citep{brown2020language} and cross-lingual classification \citep{xue-etal-2021-mt5}. However, the sometimes steep increase in performance of these tasks has led to saturation of popular benchmarks, such as MMLU \citep{hendrycks2021measuring}, where SotA performance has gone from 60\% in December 2021 \citep{rae2022scaling} to 90\% in December 2023 \citep{geminiteam2024gemini}, providing diminishing returns when it comes to quantifying differences between models.

Moreover, in the case of linguistic reasoning, the task of evaluating a model's linguistic skills is often tied to the comprehensive knowledge a model has of a certain language (most commonly, English), making it difficult to evaluate a model's underlying linguistic skills beyond language-specific knowledge.

\begin{figure}[ht]
  \centering
  
  \caption{A subset of the context of a problem in Terenâ language and the reasoning steps needed to solve it. To correctly answer the question, the model must notice that \textcolor{blue}{(a) voiced \textit{d} mutates to voiceless paired sound \textit{t} (fortition)}, \textcolor{darkgreen}{(b) \textit{n} is dropped because there are no voiceless nasal alveolar sounds} and \textcolor{alizarin}{(c) an epenthetic vowel has to be added between the mutation consonant and the rest of the word (a root), and that the vowel that gets added matches the aperture of the vowel in the root. If the aperture is closed, the epenthetic vowel is the closed front vowel \textit{i}; if the aperture is mid, the epenthetic vowel is the mid front vowel \textit{e}}.}
  \includegraphics[width=0.4\textwidth]{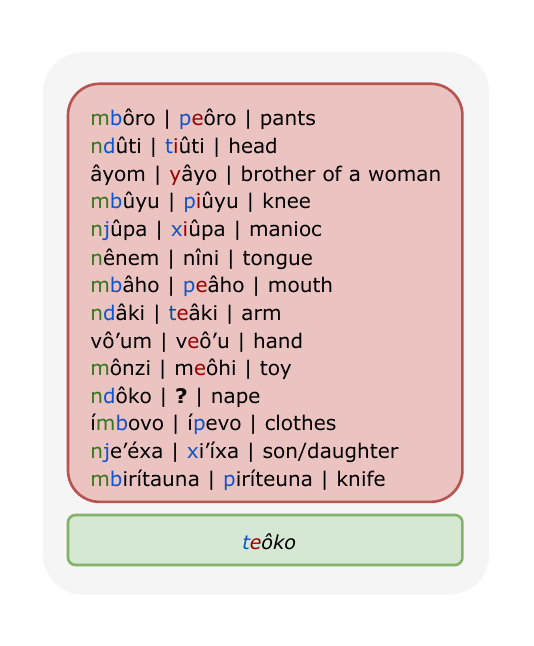} 
  
  \label{fig:roseta_stone_example}
\end{figure}

To address these issues, we introduce Linguini\footnote{The dataset is available at \url{https://github.com/facebookresearch/linguini}}, a linguistic reasoning benchmark. Linguini consists of linguistic problems which require meta-linguistic awareness and deductive reasoning capabilities to be solved instead of pre-existing language proficiency. Linguini is based on problems extracted from the International Linguistic Olympiad (IOL)\footnote{The problems are shared only for research purposes under the license CC-BY-SA 4.0. The problems are copyrighted by ©2003-2024 International Linguistics Olympiad}, a secondary school level contest where participants compete in solving Rosetta Stone-style problems \citep{derzhanski2010linguistics} relying solely on their understanding of linguistic concepts. An example of the type of challenges and the reasoning steps needs to solve it can be seen in Figure \ref{fig:roseta_stone_example}.

We evaluate a list of open and proprietary models on Linguini, showing a noticeable gap between open and closed language models, in favor of the latter. We also conduct a series of experiments aiming at understanding the role of the contextual information in the accuracy obtained in the benchmark, performing both form (transliteration) and content (removing context) ablations, with results showing a main reliance of the context to solve the problems, minimizing the impact of language or task contamination in the models' training sets.

\section{Related Work}

There has been an increasing number of articles focusing on evaluating reasoning in language models \citep{chang2023survey}. In the area of mathematical reasoning, \cite{qin2023chatgpt} analyze models' arithmetic reasoning, while \cite{frieder2023mathematical} leverage publicly-available problems to build GHOSTS, a comprehensive mathematical benchmark in natural language. \cite{bang2023multitask} include symbolic reasoning in their multitask, multilingual and multimodal evaluation suite. \cite{wu2024reasoning} and \cite{hartmann2023political} show that current language models have profound limitations when performing abstract reasoning, but \cite{liu2023evaluating} indicate promising logical reasoning skills; however, performance is limited on out-of-distribution data. Multi-step reasoning is assessed by Chain-of-Thought Hub \citep{fu2023chainofthought} and ThoughtSource \citep{Ott_2023}, pointing out the limitations of language models in complex reasoning tasks.

Coverage of linguistic reasoning, which can be defined as the ability to understand and operate under the rules of language, has been limited in evaluation datasets for language models. One of the earliest examples is PuzzLing Machines \citep{sahin-etal-2020-puzzling}, which presents 7 different patterns from the Rosetta Stone paradigm \cite{bozhanov-derzhanski-2013-rosetta} for models to perform exclusively machine translation. \cite{chi-etal-2024-modeling} replicate \cite{sahin-etal-2020-puzzling}'s approach, manually creating a number of examples to avoid data leakage. 
Recently, some approaches have leveraged long context capabilities of language models to include in-context linguistic information (e.g. a grammar book \citep{tanzer2024benchmark} and other domain-specific sources \citep{zhang2024hire}) to solve different linguistic tasks. For large-scale linguistic reasoning evaluation,  Big-Bench \citep{lewkowycz2022beyond} includes a task linguistic mappings\footnote{\url{https://github.com/google/BIG-bench/blob/main/bigbench/benchmark_tasks/linguistic_mappings/}}, relying on arbitrary artificial grammars to perform logical deduction. This approach is limited by its reliance on constructed languages instead of natural languages, which overlooks more complex underlying properties of languages, such as voicing rules. Finally, \cite{waldis2024holmes} present Holmes, a comprehensive benchmark for linguistic competence in English language.

\section{Benchmarking linguistic reasoning}

To overcome the previous limitations, we built a dataset where, in most cases, a model has no information about task language outside of the given context. To achieve this, we worked with problems extracted from the International Linguistic Olympiad.

\subsection{IOL} The International Linguistic Olympiad (IOL)\footnote{\url{https://ioling.org}} is a contest for students up to secondary school level, where contestants must compete solving problems based on their understanding of linguistics \citep{derzhanski2010linguistics}. The presented problems are formulated following the Rosetta Stone paradigm and present participants with challenges related to a variety of (mainly) extremely low-resource languages that students are not expected to be familiar with. The goal is for participants to leverage their linguistic skills rather than their foreign language knowledge. The IOL has been held yearly since 2003 (with the exception of 2020), and every year includes 5 short problems (to be solved individually) and 1 long, multipart problem (to be solved in groups). Problems are formulated in English and in several languages (up to 25 languages for the 2023 edition). The IOL corpus is available on their website in different formats of PDF with questions and correct answers, explanations of some answers and total marks for each problem. Beyond IOL, there are regional contests (e.g. Asia Pacific Linguistic Olympiad\footnote{\url{https://aplo.asia}} and The Australian Computational and Linguistics Olympiad\footnote{\url{https://ozclo.org.au}}) that award places for the IOL.

\subsection{Selecting problems for our benchmark}
\begin{figure}[htbp]
  \centering
  \caption{Examples of Linguini entries covering the three problems included in the dataset: sequence transduction, fill-in-blanks, number transliteration.}
  \includegraphics[width=0.8\linewidth]{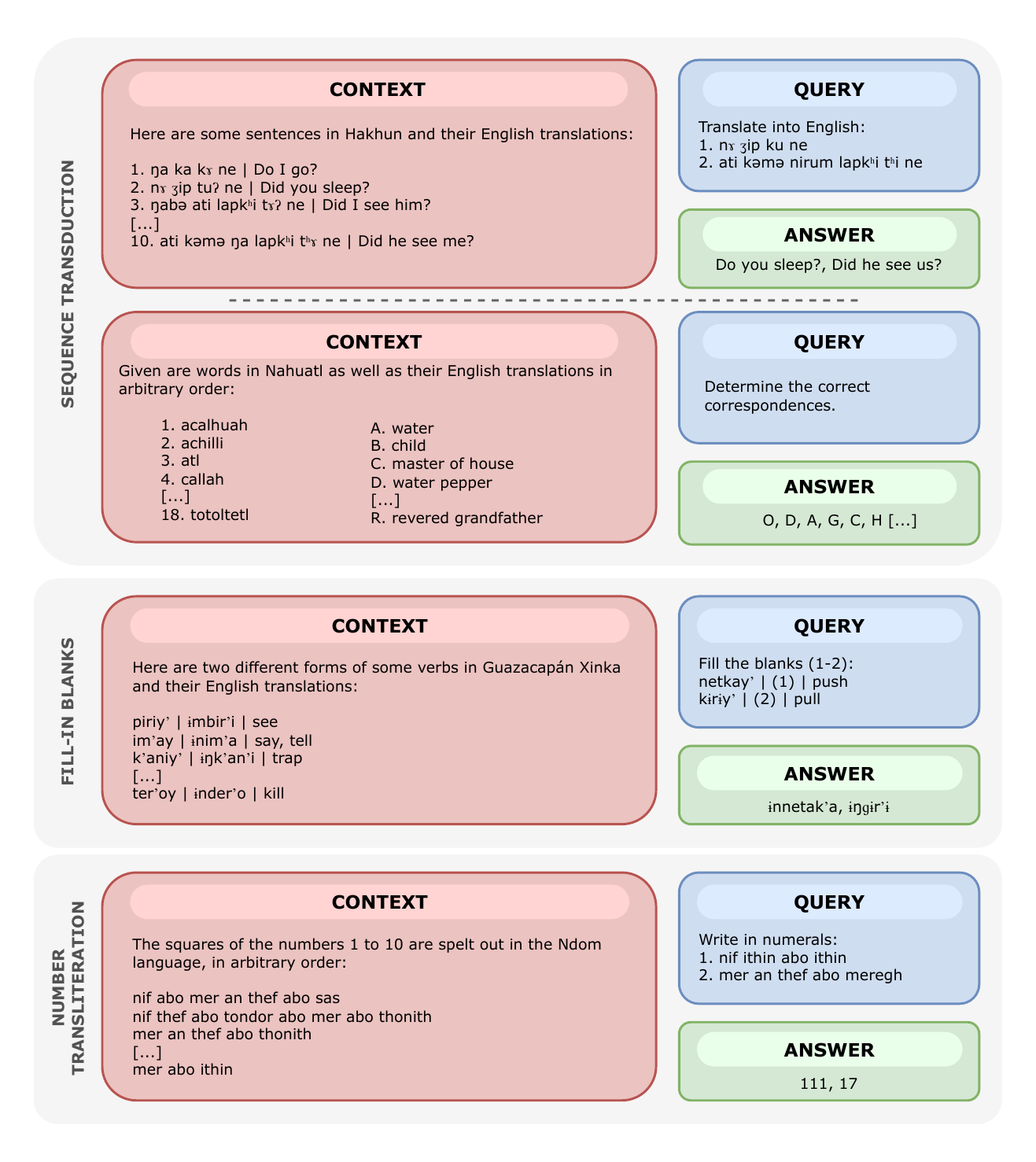}
  
  \label{fig:example_linguini}
\end{figure}

To select the types of questions for the dataset, we built a taxonomy exploring the IOL from 2003 to 2023. We excluded all instances for which their category only appears once; those where the question includes an image or those where the response is only an explanation. The remaining problems require solving different linguistic reasoning tasks, such as morphosyntactic segmentation (eg., verb conjugation), morphosemantic alignment (e.g., noun negation), derivation (e.g., finding cognates in related languages), morphophonological segmentation (e.g., pluralization) or graphophonemic transcription (e.g., transcription from one script to another). In total, Linguini is composed by 894 questions grouped in 160 problems across 75 (mostly) extremely low-resource language. A list of languages can be found in Appendix \ref{app:langs}. We classify the problems included in Linguini into the three categories according to their content: sequence transduction, fill-in-blanks and number transliteration. Figure \ref{fig:example_linguini} shows one example of each.

\paragraph{Sequence transduction} This category includes sequence production (identified in the benchmark as \verb|`translation'|) and sequence matching (identified as \verb|`match_letter'|). The problems require the model to transform a sequence into a different space (e.g., language, phonetic representation, script) based on few examples.
In some cases, basic phonetic/phonological knowledge is needed. For example, the model should be able to reason over principles of voicing and their implementation in situations of coarticulation. Some problems require to know that consonants come in voiced-voiceless pairs, and that one element of the pair may in some cases be a substitute for the other element in the pair under certain circumstances.

\paragraph{Fill-in blanks}

Fill-in blanks are mainly morphophonological derivation tasks, and they are identified in the benchmark as \verb|`fill_blanks'|.  Models need to understand what are the morphophonological rules that make it possible to go from the first form of a word to its second form. This can usually be applied to verbal (e.g., verb tense conjugation), nominal or adjectival (e.g., case declension) derivation. It involves understanding affixation rules and morpheme swapping rules, which often come with phonological rules if there are different coarticulation phenomena with different affixes or phonotactic phenomena such as consonantal mutations.

\paragraph{Digit/text number transliteration}

These problems are identified by the labels \verb|`text_to_num'| and \verb|`num_to_text'|. In them, models have to produce a digit or text equivalent, respectively. They require a model's understanding of morphological analysis and morpheme order.

\section{Experiments}
\label{sec:experiments}

We perform zero-shot to few-shot (0-5 in-context examples) evaluation across the whole dataset for an array of open and proprietary LLMs. Given the size of the benchmark, we employ a leave-one-out cross-validation scheme to maximize the number of in-context candidates per task. For every given inference, we include examples of the same format (e.g., \verb|`translation'|, \verb|`match_letter'|), but we exclude in-content examples of the same language to avoid language contamination.

\paragraph{Setup and Models}
We prompt models with an instruction, a context that provides information to unambiguously solve the linguistic problem and the problem itself. Scores of answers to each item of a problem are averaged to provide a single score (0-100) per task.  We evaluate several major open LLMs and commercially available (behind API) SotA LLMs at the publication of this work. For open models, we conduct inference experiments in an 8 A100 GPUs node. An exhaustive list can be found in Appendix \ref{app:models}.

\paragraph{Evaluation}
We use exact match (accuracy) as main evaluation criterion. Given the almost null performance on exact match of certain models, we also include chrF \citep{popovic-2015-chrf} as a \textit{softer} metric. A low ChrF score indicates extremely low performance models, e.g. not understanding the domain of the task at hand.

\section{Results and Discussion}
\label{sec:results}

Table \ref{tab:results} shows there's a gap between the best performing open model and the best performing proprietary model, with several tiers of proprietary models above the best open model (\textit{llama-3-70b}). We also find mixed impact of in-context examples in the performance of the models. While some models benefit from it (such as \textit{llama-3-70b-it}), other models' performance degrades as the number of examples increases (such as \textit{claude-3-opus}). This disparity might be due to the two factors introduced by the ICEs: from one side, they set an answer format that could be useful for models that can't infer it directly from a single natural language instruction and, from another side, they introduce tokens of languages potentially unrelated to the evaluated problem. It is possible that for models more capable of instruction following, only the second factor plays a role in the model's performance. We include results with chrF in Appendix \ref{app:chrf} for reference.

\begin{table}[ht!]
\centering
\caption{Exact match results with Linguini for 0-5 ICEs.}
\label{tab:results}
\begin{tabular}{@{}lrrrrrrr@{}}
\toprule
Model               & 0     & 1    & 2    & 3    & 4  & 5 & Best($\uparrow$)\\ \midrule
claude-3-opus       & 24.05	&20.58	&21.36	&19.91	&17.00	&15.1 &  24.05\\
gpt-4o		&	14.65&  12.98  &  13.87  &    12.98  &   13.98    & 					13.76&  14.65\\
gpt-4				&    6.38   &   9.96    &    11.52   &       12.98 &   11.74    & 13.31& 12.98\\
claude-3-sonnet		&12.30&	8.95	&10.29	&10.40	&9.28	& 8.72 & 12.30\\
gpt-4-turbo			&    8.72    &  9.40   &  9.96   &  7.49   &  8.61   &    9.96& 9.96 \\
llama-3-70b		&8.17	&5.93	&7.72	&8.84	&8.72	&6.60 & 8.84\\
llama-3-70b-it	&	4.81	&5.93	&7.16	&7.38	&6.82	&8.39 & 8.39\\
claude-3-haiku	&	6.04	&7.61	&4.36	&6.04	&6.94	& 7.05	& 7.61\\
llama-2-70b		&4.70	&2.24	&2.57	&3.24	&3.36	&3.58& 3.58\\
mistral-0.1-8x7b	&2.46	&3.47	&3.91	&3.02	&3.24	&3.47& 3.91\\
llama-2-70b-it		&0.89	&1.45	&2.80	&3.02	&3.13	&2.80& 3.13\\
gemma-2b		&0.34	&2.01	&1.90	&1.34	&1.45	&1.90& 2.01\\
qwen-1.5-110b-it	&1.45	&1.23	&1.34	&1.45	&1.45	&1.68& 1.68\\
\bottomrule
\end{tabular}
\end{table}

In addition to our main experiments, we performed a series of ablation studies to get a better insight of how language models perform linguistic reasoning.

\subsection{No-Context Prompting}

Given that we don't have information about training data for the majority of the analyzed models, we performed a series of experiments to study the degree in which models rely on the given context to provide correct answers. Models that have not been trained on any data of the task language should have a null-adjacent performance when not given the context necessary to solve the task. We analyze the impact of ignoring the context provided in the benchmark as a proxy of possible data contamination. The results are shown in Table \ref{tab:no_context}.

\begin{table}[ht!]
\centering
\caption{No context results.}
\label{tab:no_context}
\begin{tabular}{@{}lrrr@{}}
\toprule
Model               & Zero-shot     & No context  & $\Delta$  \\ \midrule
llama-3-70b-it		& 4.81	& 1.12	&  -3.69	\\ 
gpt-4-turbo			&	8.72	&	1.45	& -7.27 \\ 
gpt-4				&	6.38	&	1.34	& -5.04	\\ 
claude-3-sonnet		& 12.30	& 2.01	& -10.29	\\ 
mistral-0.1-8x7b	& 2.46	& 1.98	&  -0.48	\\ 
claude-3-haiku		& 6.04	&1.12	&  -4.92	\\ 
qwen-1.5-110b-it	& 1.45	& 0.43	&  -1.02	\\ 
gemma-2b		& 0.34	& 0.09	& -0.25	\\ 
llama-2-70b		& 4.70	& 1.07	&  -3.63	\\ 
llama-2-70b-it		& 0.89	& 0.56	&  -0.33	\\ 
llama-3-70b		& 8.17	& 1.67	& -6.50	\\ 
claude-3-opus		& 24.05	&1.23	&  -22.82	\\ 
gpt-4o			&14.65	&1.45	& -13.20 \\ 
\bottomrule
\end{tabular}
\end{table}

We find steep performance drops for every model, which points towards a low likelihood of the language (or the training examples) being present in the models' training sets.

\subsection{Character-wise substitution}

\begin{figure}[ht]
  \centering
  
  \caption{Example of transliteration of a problem into Cyrillic, Greek, Georgian and Armenian scripts.}
  \includegraphics[width=0.79\textwidth]{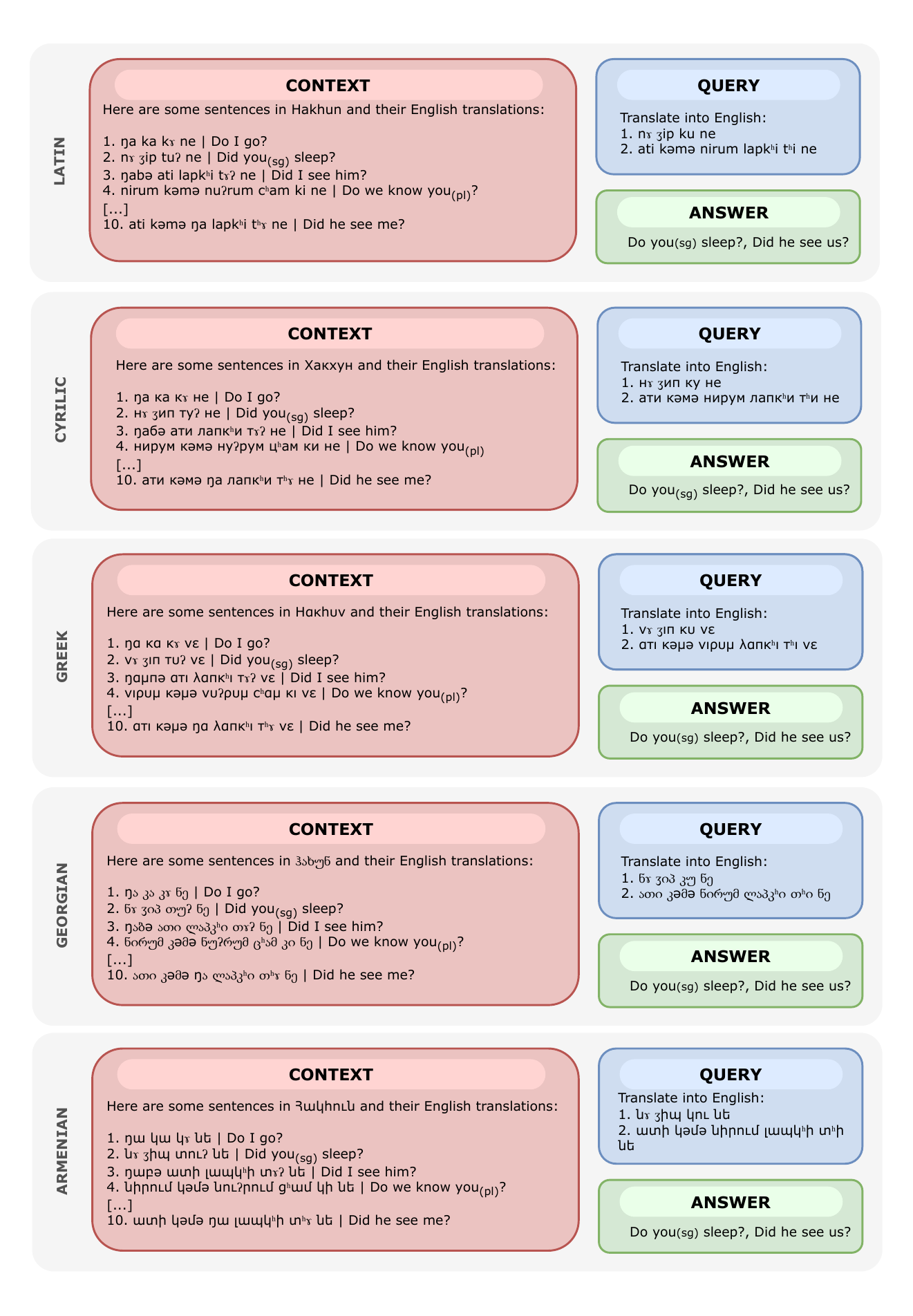}
  \label{fig:transliteration}
\end{figure}

Since most problems are presented in Latin script, we wanted to understand whether the script in which the task languages are presented impact the performance on Linguini. But given that all information needed to solve the task is present in the context, the script should not have a major impact on the performance beyond encoding constraints. In other words, if the model doesn't rely on instances of the language (or the problem) in its training set, it should be able to solve the task in a non-Latin script as well. We selected the best performing model (\textit{claude-3-opus}) and transcribed the best performing problems (those where the accuracy >= 75) into 4 non-Latin alphabetical scripts (Cyrilic, Greek, Georgian and Armenian)\footnote{The mappings from Latin script to the rest can be found at \url{https://github.com/barseghyanartur/transliterate/}}. An example of a transliterated problem can be found in Figure \ref{fig:transliteration}.
Given the difficulty of uniformly transcribing a diverse set of orthographic systems and diacritics, we opted for performing a character/bi-character-wise substitution of the standard Latin alphabet character, leaving non-standard characters with their original Unicode symbol. We filtered 17 well performing problems, and excluded one with a non-Latin script task language (English Braille). We performed transcriptions on the remaining 16 problems.

\begin{table}[ht]
\centering
\caption{Scores of selected problems with different language scripts for \textit{claude-3-opus}.}
\begin{tabular}{lrrrrr}
\toprule
\textbf{Problem code \& language} & \textbf{Latn} & \textbf{Cyrl} & \textbf{Grek} & \textbf{Geor} & \textbf{Armn} \\
\hline
012023010100 (qda-gua) & 75.00&	100.00	&75.00	&100.00	&0.00 \\
012021020500 (zun)	&100.00&	0.00&	100.00&	0.00&	0.00\\
012012030100 (eus)	&78.57&	7.14	&92.86	&0.00	&0.00\\
012018020100 (nst-hkn)	&83.33&	83.33&	66.67&	83.33	&100.00\\
012007050100 (tur)&	75.00&	75.00&	50.00	&37.50&	50.00\\
012006020100 (cat)&	75.00	&50.00&	50.00&	58.33&	33.33\\
012003030200 (eus)&	100.00&	100.00&	75.00&	100.00&	100.00\\
012004010100 (txu)&	100.00&	100.00&	66.67&	66.67&	33.33\\
012007030100 (kat)&	80.00&	13.33&	6.67&	100.00	&0.00\\
012009050100 (nci)&	83.33&	83.33&	83.33	&83.33	&50.00\\
012015020100 (kbd-bes)	&100.00&	66.67&	100.00	&66.67&	83.33\\
012012050100 (rtm)	&100.00&	100.00&	100.00	&100.00&	100.00\\
012011040200 (nci)&	100.00	&50.00&	75.00	&75.00	&0.00\\
012013010200 (yii)&	100.00&	100.00	&100.00	&75.00&	100.00\\
012012030200 (eus)&	100.00	&50.00&	0.00	&0.00&	0.00\\
012012030300 (eus)&	100.00&	50.00&	100.00&	0.00&	0.00 \\
\hline
Average                 & 85.71 & 56.12 & 65.31 & 63.27 & 38.78 \\
\bottomrule
\end{tabular}

\label{tab:scripts}
\end{table}

Table \ref{tab:scripts} shows that the model retains the capacity to perform linguistic reasoning even after changing scripts, which backs the hypothesis of the model relying mainly on the presented context and not on spurious previous knowledge. The fact that for 13 our of 16 of the given problems there's at least one non-Latin script in which the model can solve the problem with greater or equal performance than with Latin script further supports this claim.
Performance disparity among scripts could be related to either the difference in tokenization of different scripts or to the inherent limitations of our transliteration strategy (e.g. the Armenian script might lack a specific consonant cluster that needs to be developed to provide the right answer, and character/bi-character-wise substitution doesn't take this nuance into account).

\subsection{Language resourcefulness and accuracy}

\begin{figure}[ht]
  \centering
  
  \caption{Accuracy vs. number of speakers. Data points are clustered for readability.}
  \includegraphics[width=0.8\textwidth]{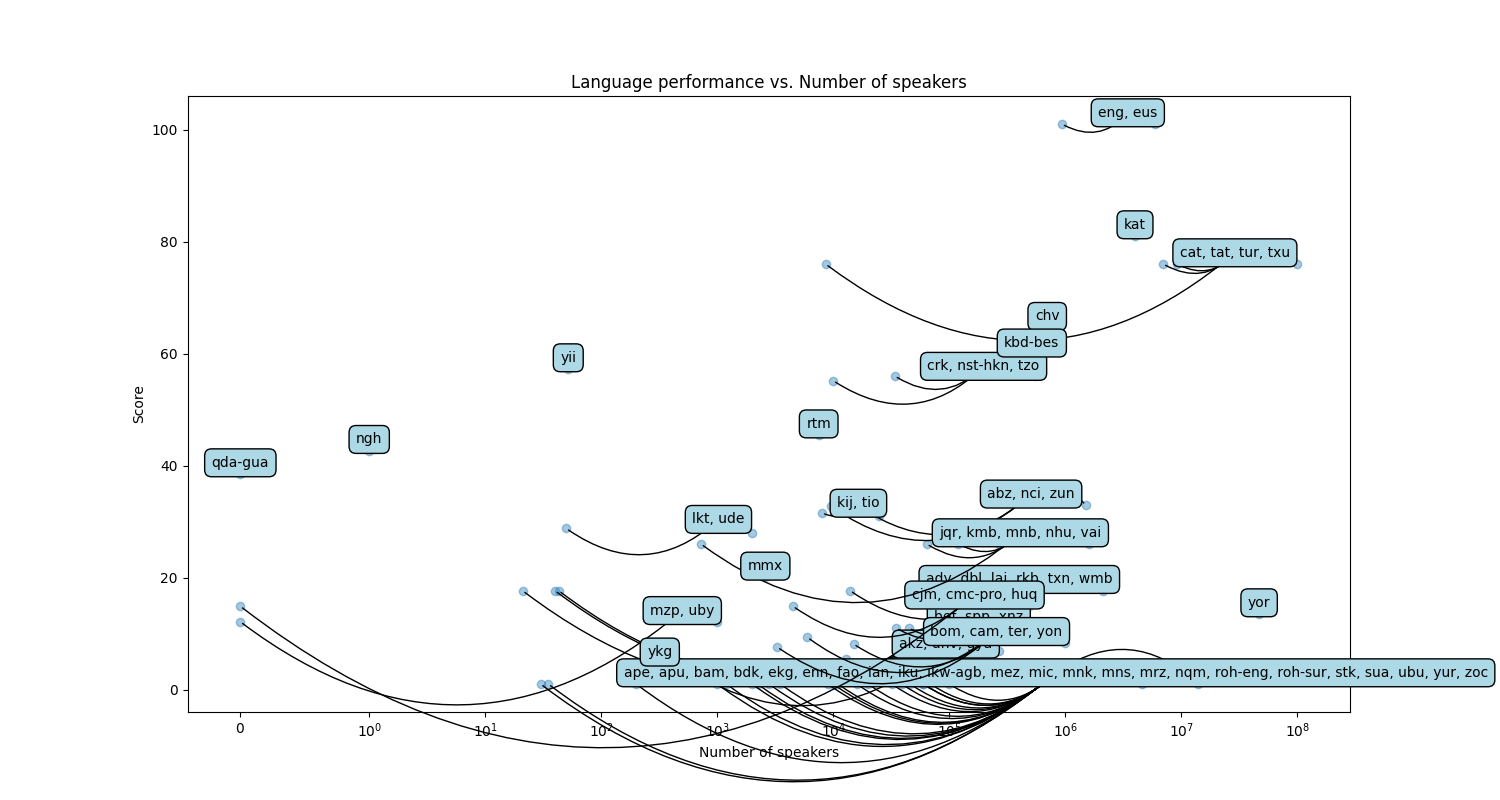}
  \label{fig:speakers}
\end{figure}

\begin{figure}[ht]
  \centering
  
  \caption{Accuracy vs. number of Google searches. Data points are clustered for readability.}
  \includegraphics[width=0.8\textwidth]{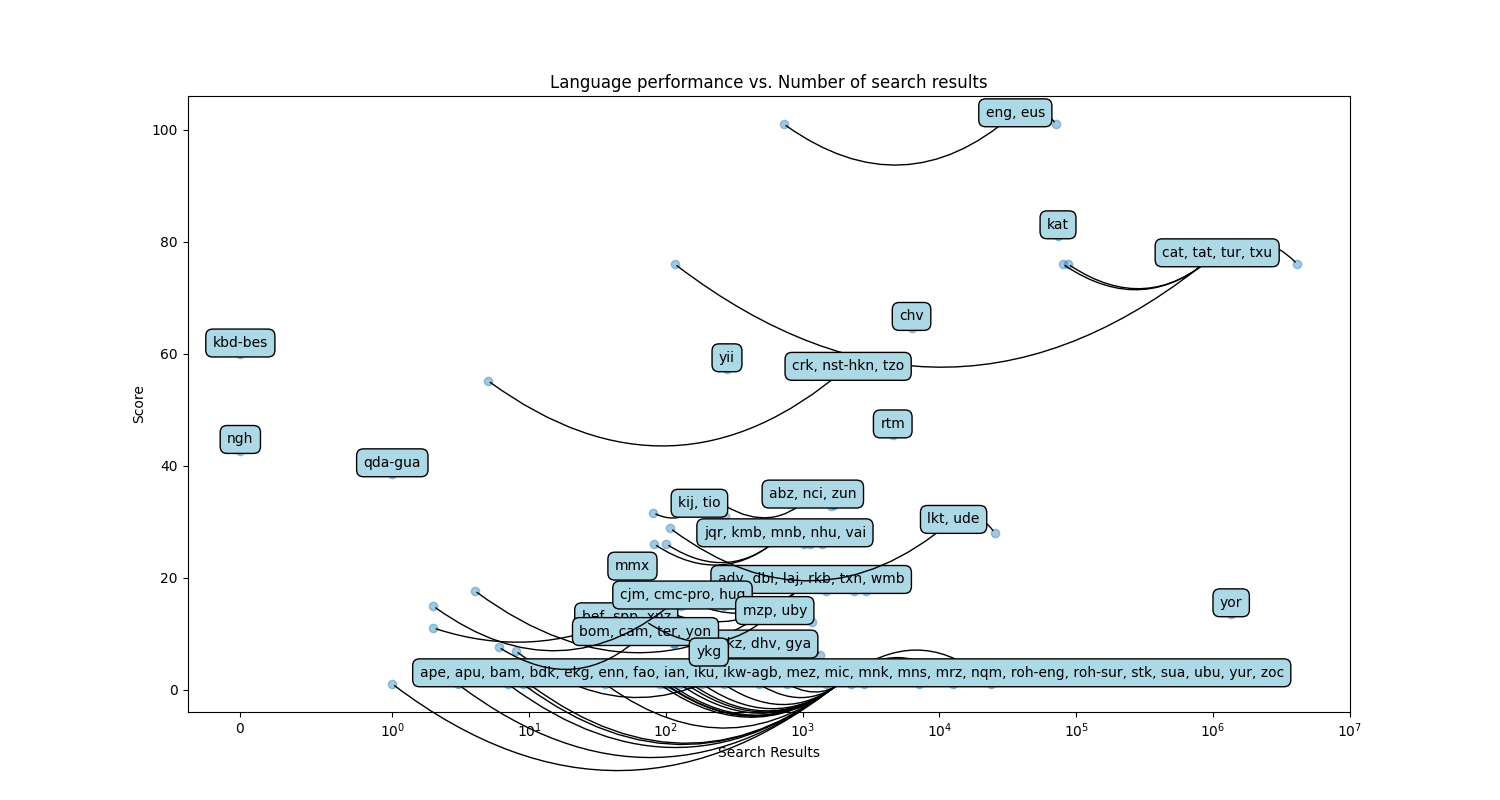}
  \label{fig:searches}
\end{figure}

We were also interested in assessing whether higher-resource languages perform, on average, better than lower-resource languages. We use two metrics as proxies of language resourcefulness: number of speakers (Figure \ref{fig:speakers}) and online presence (Figure \ref{fig:searches}), measured by Google searches).

We find the distribution to follow a uniform trend with respect to both metrics of language resourcefulness, which suggests that the accuracy isn't largely correlated to to its likelihood of being included in the training set. Notable exceptions to this trend are a number of very high-resource languages (e.g., cat, eus, kat, tur), which are very likely to be included in the model's training set, given their institutional status. 

\subsection{One-Book Prompting}
Previous studies \citep{tanzer2024benchmark} have shown the capacity of language models to acquire some proficiency in the task of machine translation for an unseen language only through an in-context textbook. We leverage publicly available textbooks to scale \citet{tanzer2024benchmark}'s analysis in number of languages and types of tasks. We convert the textbooks in PDF format to raw text using the pdftotext library\footnote{\url{https://github.com/jalan/pdftotext}} and include them as context without any pre-processing. A list of textbooks employed can be found in Appendix \ref{app:books}.

\begin{table}[ht]
\centering
\caption{Scores for a subset of examples evaluated with no context, with context, with a textbook and with a combination of both.}
\begin{tabular}{lrrrr}
\toprule
\textbf{Language code} & \textbf{No-context} & \textbf{Context} & \textbf{Textbook} & \textbf{Context + Textbook} \\
\hline
akz & 0.00 & 5.13 & 0.00 & 3.85\\
apu & 0.00 & 0.00 & 0.00 & 16.67\\
mnk & 0.00 & 0.00 & 0.00 & 0.00 \\
\hline
Average                 & 0.00 & 1.71 & 0.00 & 6.84\\
\bottomrule
\end{tabular}

\label{tab:books}
\end{table}

\begin{figure}[ht]
  \centering
  \caption{Example of transliteration of a problem into Cyrillic, Greek, Georgian and Armenian scripts. The discrepancies between the term \textit{kyky} (English: \textit{man}) in the original document (a scan from a 1894 grammar book of Apurinã language), its OCR conversion and the text of a problem in the benchmark are highlighted. In spite of the noise introduced by different orthographies and imperfect OCR, performance for Apurinã increases from 0\% 16.67\% with the full OCR text in-context.}
  \includegraphics[width=\textwidth]{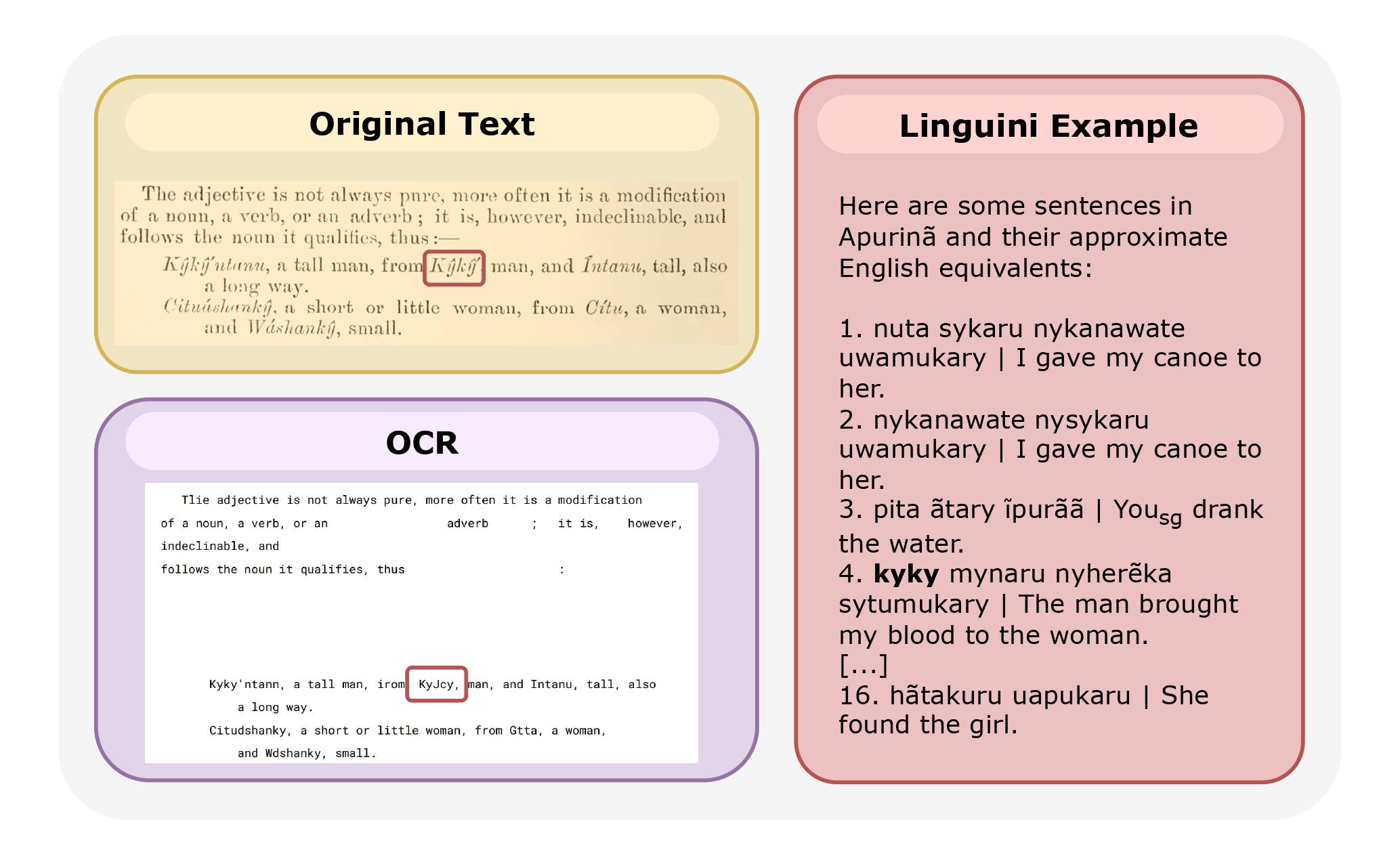}
  \label{fig:ocr}
\end{figure}

Even thought in many cases the orthography of the task language greatly varies from the textbook to the problem and the PDF to text conversion introduces errors for highly diacritical text (as shown in Figure \ref{fig:ocr}), the results in Table \ref{tab:books} show that a model can learn to model linguistic phenomena relying on a single in-context textbook.

\section{Conclusions}

We presented Linguini, a new linguistic reasoning evaluation dataset. 
Our experiments show that Linguini provides a compact and effective benchmark to assess linguistic reasoning without relying on a substrate of existing language-specific knowledge. 
There's a considerable gap between open source and proprietary LLMs in linguistic reasoning. Subsequent experiments also show very low likelihood of dataset contamination in the analyzed models. Limitations and broader impact of the dataset are discussed in Appendix \ref{app:limitations}.

\clearpage
\newpage

\bibliography{references}
\bibliographystyle{acl_natbib}

\clearpage
\newpage


\appendix

\section{Limitations, further work and broader impact}
\label{app:limitations}

Evaluation of long in-context learning for linguistic reasoning is limited in this paper to a few languages, given the difficulties of finding publicly available grammar books. We plan to scale up the number of covered languages in further versions of the benchmark to perform a better encompassing analysis of long in-context learning.

Our dataset also lacks a curated list of explanations for each problem, which could be used as a basis to run chain-of-thought experiments and improve lingusitic reasoning skills of language models. We intend to engage with linguists and IOL organizers to fill this gap. 

This benchmark intends to address and quantify the root of multilingualism, which in turn can impact the support of language models for the majority of world languages.

\section{Languages of Linguini}
\label{app:langs}

\tiny
\begin{longtable}{ccrrcc}

\caption{Languages and their characteristics} \\
\hline

\textbf{Lang. Code} & \textbf{Language} & \textbf{No. Speakers}\footnote{According to \cite{eberhardethnologue}} & \textbf{No. Search Results}\footnote{Number of search results of the exact string ''<Language name> language'' using Google Seach API} & \textbf{Language Family} & \textbf{Script}  \\ 
\hline
\endfirsthead

\hline
\textbf{Lang. Code} & \textbf{Language} & \textbf{No. Speakers} & \textbf{No. Search Results} & \textbf{Language Family} & \textbf{Script}  \\
\hline
\endhead

\hline
\endfoot

\hline
\endlastfoot
\renewcommand{\thefootnote}{\fnsymbol{footnote}}

abz & Abui & 16,000 & 263 & Trans-New Guinea & Latin\\ 
ady & Adyghe & 425,000 & 2,370 & Abkhaz-Adyghe & Latin\\
akz & Alabama & 370  & 1,350 & Muskogean & Latin\\ 
abz & Mountain Arapesh & 16,000 & 98 & Torricelli & Latin\\ 
apu & Apurinã & 2800 & 264 & Maipurean & Latin\\ 
bam & Bambara & 14000000 & 7150 & Niger-Congo & N'Ko\\ 
bdk & Budukh & 200 & 126 & Nakh-Daghestanian & Latin\\ 
bef & Bena Bena & 45000 & 107 & Trans-New Guinea & Latin\\ 
bom & Birom & 1000000 & 115 & Niger-Congo & Latin\\ 
cam & Cemuhî & 3300 & 6 & Austronesian & Latin\\
\hline
cat & Catalan & 9200000 & 87100 & Indo-European & Latin\\ 
chv & Chuvash & 700000 & 6260 & Turkic & Latin\\ 
cjm & Phan Rang Cham & 491448 & 2 & Austronesian & Latin\\ 
cmc-pro\footnotemark[11] & Proto-Chamic & 0 & 267 & Austronesian & Latin\\ 
crk & Plains Cree & 34000 & 5290 & Algic & Latin\\ 
dbl & Dyirbal & 21 & 2900 & Australian & Latin\\ 
dhv & Drehu & 13,000 & 216 & Austronesian & Latin\\ 
ekg & Ekari & 100000 & 141 & Trans-New Guinea & Latin\\ 
eng & English Braille & 6000000 & 728 & Indo-European & Latin\\ 
enn & Engenni & 20000 & 185 & Niger-Congo & Latin\\
\hline
eus & Basque & 936,812 & 71100 & Isolate & Latin\\ 
fao & Faroese & 69000 & 23800 & Indo-European & Latin\\ 
gya & Northwest Gbaya & 267000 & 8 & - & Latin\\ 
huq & Tsat & 4500 & 128 & Austronesian & Latin\\ 
ian & Iatmül & 46000 & 9 & Papua New Guinea & Latin\\ 
iku & Inuktitut & 39,000 & 12500 & Eskimo-Aleut & Latin\\ 
ikw-agb\footnotemark[11] & Agbirigba & 30 & 1 & Niger-Congo & Latin\\ 
jqr & Jaqaru & 725 & 101 & Aymaran & Latin\\ 
kat & Georgian & 4000000 & 73700 & Kartvelian & Latin\\ 
kbd-bes\footnotemark[11] & Besleney Kabardian & 516000 & 0 & Abkhaz-Adyghe & Latin\\ 
\hline
kij & Kilivila & 25000 & 271 & Austronesian & Latin\\
kmb & Kimbundu & 1600000 & 1130 & Niger-Congo & Latin\\ 
laj & Lango & 2100000 & 1490 & Nilo-Saharan & Latin\\ 
lkt & Lakhota & 2000 & 25300 & Siouan-Catawban & Latin\\ 
mez & Menominee & 2000 & 2240 & Algic & Latin\\ 
mic & Micmac & 11000 & 774 & Algic & Latin\\ 
mmx & Madak & 2600 & 57 & Austronesian & Latin\\ 
mnb & Muna & 270000 & 1020 &  Austronesian & Latin\\ 
mnk & Maninka & 4600000 & 478 & Niger-Congo & N'Ko\\ 
mns & Mansi & 2229 & 1490 & Uralic & Latin\\ 
\hline
mrz & Coastal Marind & 9000 & 100 & Trans-New Guinea & Latin\\ 
mzp & Movima & 1000 & 72 & Isolate & Latin\\ 
nci & Classical Nahuatl & 1500000 & 1690 & Uto-Aztecan & Latin\\ 
ngh & N\textbar uuki & 1 & 0 & Tuu & Latin\\ 
nhu & Nooni & 64000 & 82 & Niger-Congo & Latin\\ 
nqm & Ndom & 1200 & 154 & Trans-New Guinea & Latin\\ 
nst-hkn\footnotemark[11] & Hakhun & 10000 & 5 & Sino-Tibetan & Latin\\ 
qda-gua\footnotemark[11] & Guazacapán Xinka & 0 & 1 & Xincan & Latin\\ 
rkb & Rikbaktsa & 40 & 54 & Isolate & Latin\\ 
roh-eng\footnotemark[10] & Engadine & 60000 & 7 & Indo-European & Latin\\ 
\hline
roh-sur\footnotemark[11] & Sursilvan & 60000 & 3 & Indo-European & Latin\\ 
rtm & Rotuman & 7500 & 4560 & Austronesian & Latin\\ 
spp & Supyire & 460000 & 45 & Niger-Congo & Latin\\ 
stk & Arammba & 1000 & 36 & South-Central Papuan & Latin\\ 
sua & Sulka & 3500 & 107 & Isolate & Latin\\ 
tat & Tatar & 7000000 & 79700 & Turkic & Latin\\ 
ter & Terêna & 15,000 & 115 & Maipurean & Latin\\ 
tio & Teop & 8000 & 81 & Austronesian & Latin\\ 
tur & Turkish & 100000000 & 4130000 & Turkic & Latin\\ 
txn & West Tarangan & 14,000 & 4 & Austronesian & Latin\\ 
\hline
txu & Kayapo & 8600 & 116 & Jean & Latin\\ 
tzo & Tzotzil & 550000 & 1160 & Mayan & Latin\\ 
ubu & Umbu-Ungu & 32,000 & 90 & Trans-New Guinea & Latin\\ 
uby & Ubykh & 0 & 1180 & Abkhaz-Adyghe & Latin\\ 
ude & Udihe & 50 & 108 & Tungusic & Latin\\ 
vai & Vai & 120000 & 1380 & Niger-Congo & Latin\\ 
wmb & Wambaya & 43 & 112 & Australian & Latin\\ 
xnz & Kunuz Nubian & 35000 & 2 & Nilo-Saharan & Latin\\ 
yii & Yidiny & 52 & 280 & Australian & Latin\\ 
ykg & Tundra Yukaghir & 320 & 206 & Yukaghir & Latin\\ 
\hline
yon & Yonggom & 6,000 & 48 & Trans-New Guinea & Latin\\ 
yor & Yoruba & 47000000 & 1360000 & Niger-Congo & Latin\\ 
yur & Yurok & 35 & 2830 & Algic & Latin\\ 
zoc & Copainalá Zoque & 10000 & 10 & Mixe-Zoquean & Latin\\ 
zun & Zuni & 9500 & 1610 & Isolate & Latin\\ 
\label{tab:languages}
\end{longtable}

\footnotetext[11]{Language code not in ISO-639-3}

\renewcommand{\thefootnote}{\arabic{footnote}}

\section{Models}
\label{app:models}

\begin{table}[h]
\centering
\tiny
\caption{Overview of Large Language Models}
\begin{tabular}{cccccc}

\textbf{Model ID} & \textbf{API Version} & \textbf{Organization} & \textbf{Model Size}\footnotemark[12] & \textbf{Open} & Reference \\ \hline
claude-3-opus & claude-3-opus-20240229 & Anthropic & - &  \ding{55} & \cite{anthropic2024claude}\\
gpt-4o & gpt-4o-2024-05-13 &  OpenAI & - & \ding{55} & \cite{openai2024gpt4}\\
gpt-4 & gpt-4-0125-preview & OpenAI & - & \ding{55} & \cite{openai2024gpt4}\\
claude-3-sonnet & claude-3-sonnet-20240229 & Anthropic & - & \ding{55} & \cite{anthropic2024claude} \\
gpt-4-turbo & gpt-4-turbo-2024-04-09 & OpenAI & - & \ding{55} & \cite{openai2024gpt4} \\
llama-3-70b & - & Meta & 70.6 & \checkmark & \cite{llama3modelcard} \\
llama-3-70b-it & - & Meta & 70.6 & \checkmark & \cite{llama3modelcard} \\
claude-3-haiku & claude-3-haiku-20240307 & Anthropic & - & \ding{55} & \cite{anthropic2024claude} \\
llama-2-70b & - & Meta & 69.0 & \checkmark & \cite{touvron2023llama} \\
mistral-0.1-8x7b & - & Mistral & 46.7 & \checkmark & \cite{jiang2024mixtral} \\
llama-2-70b-it & - & Meta & 69.0 & \checkmark & \cite{touvron2023llama} \\
gemma-2b & - & Google & 2.5 & \checkmark & \cite{gemmateam2024gemma} \\
qwen-1.5-110b-it & - & Alibaba & 111.0 & \checkmark & \cite{bai2023qwen} \\
\hline
\end{tabular}
\label{tab:llms}
\end{table}

\footnotetext[12]{in billion parameter}

\renewcommand{\thefootnote}{\arabic{footnote}}

\section{Books}
\label{app:books}

\begin{table}[h]
\centering
\caption{Overview of Grammar Books [tba]}
\begin{tabular}{>{\raggedright}p{3cm}>{\raggedright}p{3cm}>{\raggedright\arraybackslash}p{3cm}}
\hline
 \textbf{Language} & \textbf{Book Title} & \textbf{Citation} \\ \hline
akz & The Language of the Alabama Indians &   \cite{lupardus1982language} \\ \hline
apu & A Grammar and a Vocabulary of the Ipuriná Language & \cite{polak1894grammar} \\\hline
mnk & The Structure of Faranah-Maninka & \cite{spears1965structure}\\ \hline
\end{tabular}
\label{tab:grammar_books}
\end{table}

\section{chrF Results}
\label{app:chrf}

\begin{table}[ht!]
\centering
\caption{chrF results with Linguini for 0-5 ICEs}
\label{tab:results_chrf}
\begin{tabular}{@{}lrrrrrr@{}}
\toprule
Model               & 0     & 1    & 2    & 3    & 4  & 5 \\ \midrule
llama-3-70b-it		&45.35	&42.65	&43.89	&45.99	&48.07	&51.08	\\ 
gpt-4-turbo			&	52.89	&	50.82	& 50.03 & 50.94 & 49.98 & 51.79	\\ 
gpt-4				&	44.62	&	55.05	& 58.47 & 57.36 & 57.62 &58.18\\ 
claude-3-sonnet		&54.97	&45.32	&50.91	&47.35	&46.51	&42.06\\ 
mistral-0.1-8x7b	&42.0	&34.8	&38.01	&37.57	&37.64	&37.63\\ 
claude-3-haiku		&47.74	&50.75	&41.02	&45.38	&42.32	& 41.83	\\ 
qwen-1.5-110b-it	&2.57	&0.0	&0.22	&0.78	&1.12	&2.8\\ 
gemma-2b		&33.72	&27.19	&24.62	&26.04	&27.04	&27.63	\\ 
llama-2-70b		&45.3	&35.39	&34.06	&35.54	&36.21	&36.44\\ 
llama-2-70b-it		&43.55	&41.42	&39.73	&41.42	&39.69	&39.34	\\ 
llama-3-70b		&37.25	&36.04	&41.83	&41.21	&41.92	&41.63	\\ 
claude-3-opus		&63.96	&58.26	&58.5	&53.17	&49.01	&46.55	\\ 
gpt-4o			&57.68	& 58.13 & 57.32 & 58.86 & 58.99 &				58.22	\\ 
\bottomrule
\end{tabular}
\end{table}


\newpage

\end{document}